\documentclass{article}

\usepackage{microtype}
\usepackage{graphicx}
\usepackage{subfigure}
\usepackage{booktabs} % for professional tables

\usepackage{amssymb}
\usepackage{amsmath}
\usepackage{amsthm}
\allowdisplaybreaks
\usepackage{xcolor}
\usepackage{makecell}

\usepackage{hyperref}

\usepackage[accepted]{icml2020}

\icmltitlerunning{Sub-Linear Memory: How to Make Performers SLiM}

\begin{document}

\twocolumn[
\icmltitle{Sub-Linear Memory: How to Make Performers SLiM}

% It is OKAY to include author information, even for blind
% submissions: the style file will automatically remove it for you
% unless you've provided the [accepted] option to the icml2020
% package.

% List of affiliations: The first argument should be a (short)
% identifier you will use later to specify author affiliations
% Academic affiliations should list Department, University, City, Region, Country
% Industry affiliations should list Company, City, Region, Country

% You can specify symbols, otherwise they are numbered in order.
% Ideally, you should not use this facility. Affiliations will be numbered
% in order of appearance and this is the preferred way.
\icmlsetsymbol{equal}{*}

\begin{icmlauthorlist}
\icmlauthor{Valerii Likhosherstov}{to}
\icmlauthor{Krzysztof Choromanski}{goo,co}
\icmlauthor{Jared Davis}{dm,st}
\icmlauthor{Xingyou Song}{goo}
\icmlauthor{Adrian Weller}{to,ed}
\end{icmlauthorlist}

\icmlaffiliation{to}{University of Cambridge}
\icmlaffiliation{goo}{Google Brain}
\icmlaffiliation{ed}{Alan Turing Institute}
\icmlaffiliation{dm}{DeepMind}
\icmlaffiliation{st}{Stanford University}
\icmlaffiliation{co}{Columbia University}

\icmlcorrespondingauthor{Valerii Likhosherstov}{vl304@cam.ac.uk}
%\icmlcorrespondingauthor{Eee Pppp}{ep@eden.co.uk}

% You may provide any keywords that you
% find helpful for describing your paper; these are used to populate
% the "keywords" metadata in the PDF but will not be shown in the document
\icmlkeywords{Machine Learning, ICML}

\vskip 0.3in
]

% this must go after the closing bracket ] following \twocolumn[ ...

% This command actually creates the footnote in the first column
% listing the affiliations and the copyright notice.
% The command takes one argument, which is text to display at the start of the footnote.
% The \icmlEqualContribution command is standard text for equal contribution.
% Remove it (just {}) if you do not need this facility.

\printAffiliationsAndNotice{}  % leave blank if no need to mention equal contribution
%\printAffiliationsAndNotice{\icmlEqualContribution} % otherwise use the standard text.

\begin{abstract}
The Transformer architecture has revolutionized deep learning on sequential data, becoming ubiquitous in state-of-the-art solutions for a wide variety of applications. Yet vanilla Transformers are notoriously resource-expensive, requiring $O(L^2)$ in serial time and memory as functions of input length $L$. Recent works proposed various linear self-attention mechanisms, scaling only as $O(L)$ for serial computation. We perform a thorough analysis of recent Transformer mechanisms with linear self-attention, \textit{Performers}, in terms of overall computational complexity. We observe a remarkable computational flexibility: forward and backward propagation can be performed \textbf{with no approximations} using \textbf{sublinear memory} as a function of $L$ (in addition to negligible storage for the input sequence), at a cost of greater time complexity in the parallel setting. In the extreme case, a Performer consumes \textbf{only $O(1)$ memory} during training, and still requires $O(L)$ time. This discovered time-memory tradeoff can be used for training or, due to complete backward-compatibility, for fine-tuning on a low-memory device, e.g. a smartphone or an earlier-generation GPU, thus contributing towards decentralized and democratized deep learning.
\end{abstract}

\begin{figure*}[ht!]
    \centering
    \includegraphics[width=\textwidth]{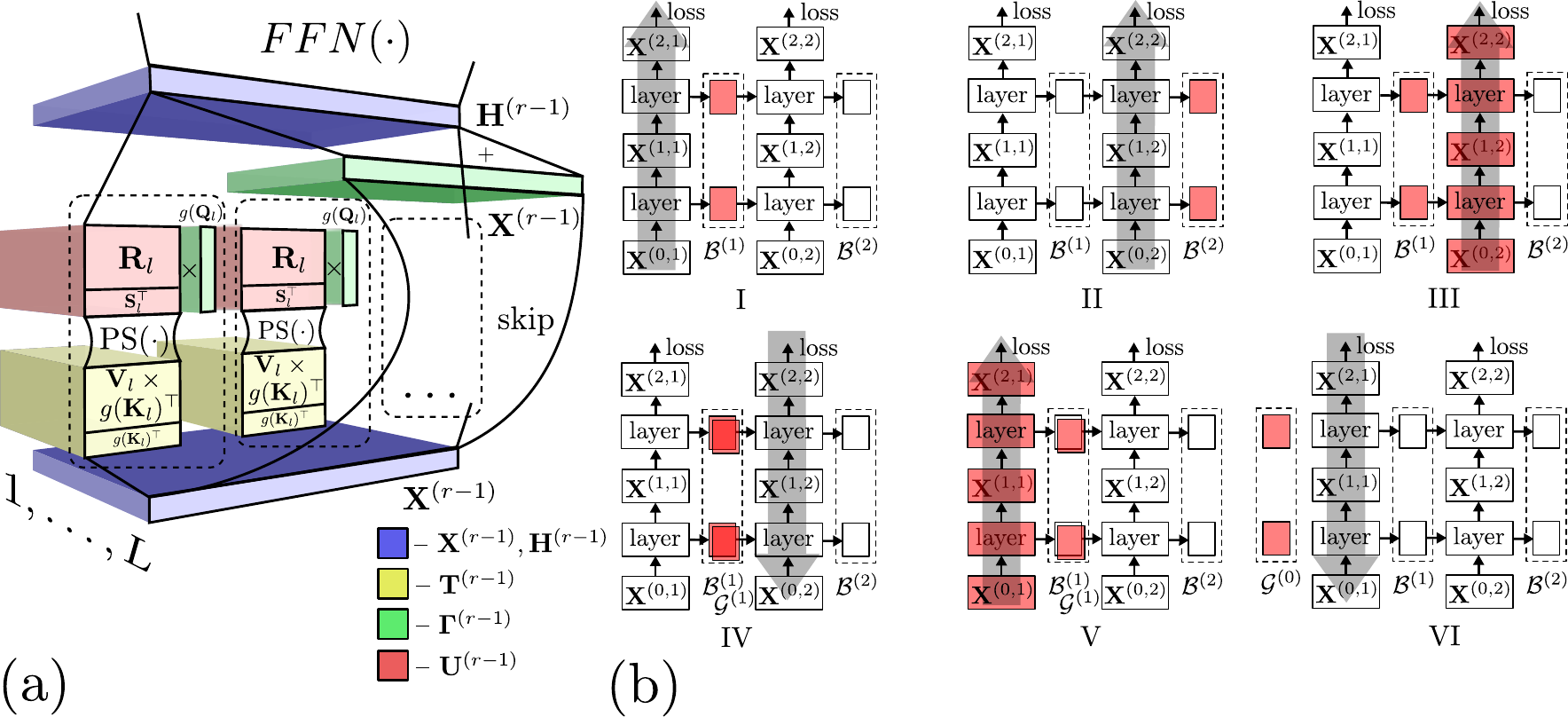}
    \caption{\textbf{(a)} $\mathrm{MultiHead}\text{-}\mathrm{Att}$ block at the $r$th layer and its decomposition into $\mathbf{T}^{(r - 1)}, \mathbf{\Gamma}^{(r - 1)}, \mathbf{U}^{(r - 1)}$. \textbf{(b)} Illustration of the Algorithm \ref{alg:1} when $r = n = 2$. I-II) forward passes for $n = 1, 2$ respectively, only the loss value and $\mathcal{B}^{(n)}$ are stored. III) backward pass start, forward computation through the slice $n = 2$ to build symbolic $\Phi^{(2)}$ and update $\mathcal{B}^{(2)} \to \mathcal{B}^{(1)}$. IV) back-propagation through $\Phi^{(2)}$ to find $\nabla_{\theta^{(2)}} \mathcal{L}$ and $\mathcal{G}^{(1)}$. V,VI) the same backward iteration for $n = 1$.}
    \label{fig:1}
\end{figure*}

\section{Introduction}

The Transformer architecture \citep{transformer} has changed the landscape of deep learning for sequential data. In contrast to more conventional methods such as recurrent neural networks \citep{lstm,gru}, the self-attention module, responsible for temporal information propagation, is fully-parallelizable, meaning that the training speed can be increased by simply using more compute resources.

However, this parallel-friendly structure of self-attention comes at a cost of quadratic $O(L^2)$ time and memory complexity, where $L$ is the length of %a sequence passed into 
the Transformer's input sequence. A recent line of work aimed to address this restriction, using either structured sparsity \citep{sparsetr}, truncated back-propagation \citep{transformerxl}, clustering \citep{reformer,routing} or linear attention methods \citep{transfrnn,performer,factatt,linattseg}. For a detailed overview of efficient Transformers, see \citep{efftran}. We refer to the family of linear attention architectures as \textit{Performers}, following \citet{performer}, since their generic kernel formulation covers all the aforementioned linear attention methods. Performers reduce time and memory complexity to linear $O(L)$ and can provably approximate conventional quadratic Transformers \citep{performer}, demonstrating strong performance in a systematic comparison of efficient Transformers \citep{lra}.

This recent trend of feeding longer sequences into Transformers, coupled with the use of deeper models, introduces new challenges for researchers and practitioners. Whereas conventional Transformer setups benefit from large-batch optimization \citep{lamb}, long sequence modelling necessitates smaller batch sizes in order to fit the model into memory. For instance, \citet{reformer} used a batch size of 1 per TPU chip to fit 64K-long sequences into their Reformer model. \citet{transfrnn} took a batch size of 4 to fit flattened CIFAR-10 images (length 3K) into their Performer analog trained on an NVidia P40 GPU with 24GB memory. \citet{performer} could use a batch of at most 8 protein sequences (length 8K, TrEMBL dataset) per TPU chip to train Performer. Aiming to use larger batch sizes, practitioners introduced various tricks. One of them, included in the popular Transformer library Fairseq \citep{fairseq} and called \textit{gradient accumulation} \citep{gradacc}, splits the batch into smaller chunks that are evaluated sequentially and then the resulting batch gradient is accumulated.

As the sequence length increases, even a batch size of 1 is too big for memory rendering training impossible. This problem is especially pronounced for low-memory devices, such as earlier-generation GPUs or smartphones. Heuristics, such as chunking the input into subsegments or truncated back-propagation \citep{transformerxl}, limit gradient propagation across the whole input, and, consequently, %do not lead to a 
impair long-context pattern learning.

We propose a solution based on the analysis of Performers. We discover a remarkable property: even for batch size of 1, a user can decrease memory consumption at the cost of smaller parallel bandwidth of the model. Notably, \textbf{no approximations are introduced, so the obtained gradient is correct and backward-compatible}. Our proposed long-sequence training algorithm can be used for training or fine-tuning on a low-memory device, thus contributing towards decentralized and democratized deep learning. The algorithm has the following advantages:
\begin{enumerate}
    \item The parameter $C, 1 \leq C \leq L,$ controls a tradeoff between the memory, scaling as $O(C)$ in addition to a negligible input sequence storage, and parallel running time ($O((L / C) \log C)$). When $C = 1$, \textbf{the algorithm consumes as much memory as if a single token were fed into Performer}, plus a small addition.
    \item The algorithm does not introduce many additional computations: for any $C$, \textbf{it requires as many floating point operations (FLOPs) as two full-memory forward and one backward passes} plus a small addition.
    \item We outline conditions when the algorithm can be extended beyond Performers. By doing so, we hope to facilitate exploration of new memory-cheap architectures to benefit deep learning more generally.
\end{enumerate}

We evaluate the proposed time-memory tradeoff empirically, and confirm backward-compatibility for language modelling on a copying task, Penn Treebank \cite{ptb} and Enwik8 \citep{enwik8} datasets.\footnote{Code: \url{https://github.com/google-research/google-research/tree/master/performer/models/slim\_performer}.}

\section{Background}

\subsection{Exponential and Linear Self-Attention} \label{sec:sa}

We commence by defining \textit{exponential self-attention} \citep{transformer}, a key component of the Transformer. Consider a sequence scale $l \in \{ 1, \dots, L \}$ and three matrices: \textit{queries} $\mathbf{Q} \in \mathbb{R}^{L \times d}$, \textit{keys} $\mathbf{K} \in \mathbb{R}^{L \times d}$ and \textit{values} $\mathbf{V} \in \mathbb{R}^{L \times d}$. Then exponential self-attention is defined as a functional producing $\mathbf{Y} = \mathrm{Att}^{exp} (\mathbf{Q}, \mathbf{K}, \mathbf{V}) \in \mathbb{R}^{L \times d}$,
\begin{equation}
    \forall l \in \{ 1, \dots, L \}: \mathbf{Y}_l = \frac{\sum_{l' = 1}^l \exp (\mathbf{Q}_l^\top \mathbf{K}_{l'}) \mathbf{V}_{l'}}{\sum_{l' = 1}^l \exp (\mathbf{Q}_l^\top \mathbf{K}_{l'})}, \label{eq:biatt}
\end{equation}
where by $\mathbf{Z}_l \in \mathbb{R}^{d_2 \times \dots}$ we denote slice $\mathbf{Z}_{l,:,\dots,:}$ of a tensor $\mathbf{Z} \in \mathbb{R}^{d_1 \times d_2 \times \dots}$. Mapping (\ref{eq:biatt}) is designed as a differentiable dictionary, where output at index $l$ is a weighted average over value vectors $\mathbf{V}_{:l}$.
For needs of autoregressive generative modelling, when each element depends only on previous elements of the sequence \citep{transformer}, $\mathbf{Y}_l$ only depends on inputs at indices $\{ 1, \dots, l \}$. Self-attention of type (\ref{eq:biatt}) is a key contributor to state-of-the-art results in many applications. However, its running time and memory scale as $O (L^2)$. This prevents applicability of exponential self-attention to sequences of big length $L \gg d$. Hence, \textit{linear self-attention} methods were proposed \citep{transfrnn,performer,factatt,linattseg}, where the exponent is substituted by a Euclidean inner-product. This is defined as a functional $\mathbf{Y} = \mathrm{Att}^{lin} (\mathbf{Q}, \mathbf{K}, \mathbf{V}) \in \mathbb{R}^{L \times d}$, where
\begin{gather}
    \forall l \in \{ 1, \dots, L \} : \mathbf{Y}_l = \frac{\sum_{l' = 1}^l \mathbf{V}_{l'} \cdot (g (\mathbf{K}_{l'})^\top g (\mathbf{Q}_l))}{\sum_{l' = 1}^l g (\mathbf{K}_{l'})^\top g (\mathbf{Q}_l)} \nonumber \\
    = \frac{(\sum_{l' = 1}^l \mathbf{V}_{l'} \times g (\mathbf{K}_{l'})^\top) \times g (\mathbf{Q}_l)}{( \sum_{l' = 1}^l g (\mathbf{K}_{l'}) )^\top g (\mathbf{Q}_l)}, \label{eq:mcatt}
\end{gather}
where ``$\times$'' denotes a matrix-matrix or matrix-vector product and $g: \mathbb{R}^d \to \mathbb{R}_+^M$ is a mapping into a vector with positive elements. The positivity of the result is to guarantee that the division in (\ref{eq:mcatt}) is well-defined and stable. In practice, $M$ is chosen to be much smaller than $L$. $g (\cdot)$ can be chosen as a simple elementwise mapping (so that $d = M$). \citet{performer} propose a randomized form of $g (\cdot)$, which is an unbiased approximation to exponential self-attention (\ref{eq:biatt}). The second transition in (\ref{eq:mcatt}), which is due to associativity of matrix multiplication, suggests an algorithm to compute linear self-attention efficiently in subqudratic time.

For a series of tensors $\mathbf{Z}^{(1)}, \dots, \mathbf{Z}^{(n)}$ of the same shape, by $\mathbf{Z} = (\mathbf{Z}^{(i)})_{i = 1}^n$ we understand a tensor such that for all $1 \leq i \leq n$ $\mathbf{Z}_{i,:,\dots,:} = \mathbf{Z}^{(i)}$. By $\mathbf{R} \in \mathbb{R}^{L \times d \times M}$, $\mathbf{S} \in \mathbb{R}^{L \times M}$ denote a tensor and a matrix such that
\begin{equation}
    \mathbf{R} = \mathrm{PS} ( ( \mathbf{V}_l \times g ( \mathbf{K}_l )^\top )_{l = 1}^L ), \,\, \mathbf{S} = \mathrm{PS} ( ( g ( \mathbf{K}_l ) )_{l = 1}^L ), \label{eq:pspart}
\end{equation}
where $\mathrm{PS} (\mathbf{Z}) = (\sum_{i' = 1}^i \mathbf{Z}_{i'})_{i = 1}^n$ is an operator taking a \textit{prefix sum} (or a \textit{cumulative sum}) along the first dimension of the input tensor $\mathbf{Z}$. Next, compute
\begin{equation}
    \forall 1 \leq l \leq L : \mathbf{Y}_l = ( \mathbf{R}_l \times g (\mathbf{Q}_l) ) / ( \mathbf{S}_l^\top g (\mathbf{Q}_l) ) . \label{eq:yappr}
\end{equation}

Depending on the prefix-sum algorithm used in (\ref{eq:pspart}), we can obtain different complexity estimates for linear self-attention. \citet{transfrnn} propose to iterate through $l = 1, \dots, L$ maintaining only current $\mathbf{R}_l, \mathbf{S}_l$, and compute and store the result $\widetilde{\mathbf{Y}}_l$. This way, tensors $\mathbf{R}, \mathrm{PS}( \mathbf{R} ) \in \mathbb{R}^{L \times d \times M}$ are not stored in memory, resulting in $O(L)$ time complexity and $O(L (d + M) + d M)$ memory complexity. \citet{transfrnn} also propose a similar iterative scheme for computing gradients through (\ref{eq:pspart}-\ref{eq:yappr}); see Appendix \ref{sec:biter} for a detailed discussion.

Alternatively, \citet{performer} employ a parallel prefix-sum algorithm \citep{cumsum,parsum}, which, for a tensor $\mathbf{Z} \in \mathbb{R}^{L \times \dots}$, finds $\mathrm{PS} (\mathbf{Z})$ in $O(\log L)$ parallel time and $O(L)$ memory. Applying this algorithm for computing $\mathrm{PS}( \mathbf{R} )$, $\mathrm{PS} (\mathbf{S} )$ and then computing (\ref{eq:yappr}) results in only $O(\log L)$ parallel time complexity and $O(L d M)$ memory consumption.

\subsection{Transformer and Performer Architectures}

In this subsection we outline a Transformer architecture which is used for autoregressive language modelling \citep{imtransf}. We focus on language modelling: first, to simplify notation, while our subsequent derivations are applicable in broader setups; second, language models are a crucial class of architectures because they were shown to act as few-shot learners, e.g. the seminal GPT-2 \citep{radford} and GPT-3 \citep{gpt-3}.

Let $\mathbf{p} \in \Sigma^L$ be an input sequence of length $L$, where $\Sigma$ is a finite alphabet. By $\mathrm{emb} (\mathbf{p}_l, l) \in \mathbb{R}^{d_{model}}$, $1 \leq l \leq L$, denote a linear combination of the $\mathbf{p}_l$ token's learned embedding and positional embedding of $l$'s position (sinusoids with different frequencies, as in \citep{transformer}). Then \textit{Transformer} is defined as a parametrized mapping from $\mathbf{X}^{(0)} = (\mathrm{emb}(\mathbf{p}_l, l))_{l = 1}^L \in \mathbb{R}^{L \times d_{model}}$ into $\mathbf{X}^{(out)} \in \mathbb{R}^{L \times | \Sigma |}$ through a sequence of hidden representations $\mathbf{X}^{(1)}, \dots, \mathbf{X}^{(s)} \in \mathbb{R}^{L \times d_{model}}$. More formally, $\mathbf{X}^{(out)} = \mathbf{X}^{(s)} \mathbf{W}^{(out)} + \mathbf{b}^{(out)}$ and for each $1 \leq r \leq s$:
\begin{gather}
    \mathbf{H}^{(r - 1)} = \mathrm{LN} (\mathrm{MultiHead}\text{-}\mathrm{Att} (\mathbf{X}^{(r - 1)})) + \mathbf{X}^{(r - 1)}, \label{eq:tr1} \\
    \mathbf{X}^{(r)} = \mathrm{LN} (\mathrm{FFN} (\mathbf{H}^{(r - 1)})) + \mathbf{H}^{(r - 1)}, \text{ where } \label{eq:tr2} \\
    \mathrm{MultiHead}\text{-}\mathrm{Att} (\overline{\mathbf{X}}) = [ \mathbf{H}^{(1)} \, \dots \, \mathbf{H}^{(k)} ], \label{eq:tr3} \\
    \forall j \leq k: \mathbf{H}^{(j)} = \mathrm{Att} (\overline{\mathbf{X}} \mathbf{W}_Q^{(j)}, \overline{\mathbf{X}} \mathbf{W}_K^{(j)}, \overline{\mathbf{X}} \mathbf{W}_V^{(j)}), \label{eq:tr4} \\
    \mathrm{FFN}( \overline{\mathbf{H}} ) = \mathrm{GeLU} (\overline{\mathbf{H}} \mathbf{W}^{(1)} + \mathbf{b}^{(1)}) \mathbf{W}^{(2)} + \mathbf{b}^{(2)} . \label{eq:tr5}
\end{gather}
Here $\mathrm{Att}$ is either $\mathrm{Att}^{exp}$ or $\mathrm{Att}^{lin}$ and $k$ is the number of attention heads ($d_{model} = k d$). $\mathbf{W}^{(out)} \in \mathbb{R}^{d_{model} \times | \Sigma |}$, $\mathbf{b}^{(out)} \in \mathbb{R}^{1 \times | \Sigma |}$, $\mathbf{W}^{(1)} \in \mathbb{R}^{d_{model} \times d_{ff}}$, $\mathbf{b}^{(1)} \in \mathbb{R}^{1 \times d_{ff}}$, $\mathbf{W}^{(2)} \in \mathbb{R}^{d_{ff} \times d_{model}}$, $\mathbf{b}^{(2)} \in \mathbb{R}^{1 \times d_{model}}$, $\mathbf{W}^{(j)}_Q, \mathbf{W}^{(j)}_K, \mathbf{W}^{(j)}_V \in \mathbb{R}^{d_{model} \times d}$ are trainable parameters (separate for each instance of $\mathrm{MultiHead}\text{-}\mathrm{Att}$, $\mathrm{FFN}$), ``$+$'' is broadcasted rowwise when biases are added and $\mathrm{LN}$ is layer normalization \citep{ln}, which is applied rowwise and depends on additional trainable parameters. $\mathrm{GeLU}$ denotes Gaussian error Linear Unit \citep{gelu}, which is applied elementwise. We refer to the Transformer (\ref{eq:tr1}-\ref{eq:tr5}) with linear self-attention $\mathrm{Att}^{lin}$ as \textit{Performer}.

For each $1 \leq l \leq L - 1$, \- $\mathbf{X}^{(out)}_l$ denotes predicted logits of the probability distribution over the next token $\mathbf{p}_{l + 1}$. Let $\mathcal{L}_l (\mathbf{X}^{(out)}_l)$ denote a cross-entropy loss with respect to $\mathbf{p}_{l + 1}$, or zero when $l = L$. The minimized loss is defined as
\begin{equation}
    \mathcal{L} = (L - 1)^{-1} \cdot (\mathcal{L}_1 (\mathbf{X}^{(out)}_1) + \dots + \mathcal{L}_L (\mathbf{X}^{(out)}_L)) . \label{eq:loss}
\end{equation}

The Transformer configuration (\ref{eq:tr1}-\ref{eq:tr5}) can be slightly changed in the literature: different $\mathrm{LN} (\cdot)$ placement, $\mathrm{GeLU}$ replaced with $\mathrm{ReLU}$, etc. The discussed variant (\ref{eq:tr1}-\ref{eq:tr5}) corresponds to GPT-2. We consider this configuration for simplicity and use it in experiments. However, as we further show, our findings can be easily extended to other modifications.

\section{Low-Memory Training Algorithm}

\subsection{Compact Notation for Performer}

In this section we consider Performer: the Transformer defined by (\ref{eq:tr1}-\ref{eq:tr5}) with $\mathrm{Att} = \mathrm{Att}^{lin}$. In light of the definition (\ref{eq:tr1}-\ref{eq:tr5}) and the algorithm for linear self-attention evaluation (\ref{eq:pspart}-\ref{eq:yappr}), the sequence of computations $\mathbf{X}^{(0)} \to \mathbf{X}^{(1)} \to \dots \to \mathbf{X}^{(s)}$ can be rewritten in the following compact form, which is more convenient for our subsequent analysis. For each $1 \leq r \leq s$,
\begin{gather}
    \mathbf{T}^{(r - 1)}, \mathbf{\Gamma}^{(r - 1)} = F^{(r)} (\mathbf{X}^{(r - 1)}; \theta), \label{eq:gtr1} \\
    \mathbf{U}^{(r - 1)} = \mathrm{PS} ( \mathbf{T}^{(r - 1)} ), \label{eq:gtr2} \\
    \mathbf{X}^{(r)} = G^{(r)} (\mathbf{U}^{(r - 1)}, \mathbf{\Gamma}^{(r - 1)} ; \theta). \label{eq:gtr3}
\end{gather}
Here $\theta \in \mathbb{R}^{n_{param}}$ is a set of all trainable parameters, $\mathbf{T}^{(r - 1)}, \mathbf{U}^{(r - 1)} \in \mathbf{R}^{L \times D_1}$ and $\mathbf{\Gamma}^{(r - 1)} \in \mathbf{R}^{L \times D_2}$ are the following matrices (see Figure \ref{fig:1}a for an illustration):
\begin{itemize}
    \item $\mathbf{T}^{(r - 1)}$ is a matrix of intermediate representations which are passed into the prefix-sum operator. That is, for each $1 \leq l \leq L$, $\mathbf{T}^{(r - 1)}_l$ is a concatenation of $g(\mathbf{K}_l)$ and flattened $\mathbf{V}_l \times g(\mathbf{K}_l)^\top$ for all attention heads computed at the $r$th step (Equations \ref{eq:tr4} and \ref{eq:pspart}). Consequently, $D_1 = M(d + 1)k$.

    \item For each $1 \leq l \leq L$, $\mathbf{U}^{(r - 1)}_l$ is a concatenation of all corresponding $\mathbf{S}_l$ and flattened $\mathbf{R}_l$ -- results of the prefix-sum operation (Equation \ref{eq:pspart}) inside each self-attention head (Equation \ref{eq:tr4}).

    \item $\mathbf{\Gamma}^{(r - 1)}$ is a matrix of representations which skip the prefix-sum operation. For each $1 \leq l \leq L$, $\mathbf{\Gamma}^{(r - 1)}_l$ is a concatenation of $\mathbf{X}^{(r - 1)}_l$ and $g( \mathbf{Q}^{(j)}_l ) = g( \overline{\mathbf{X}} \mathbf{W}_Q^{(j)} )$ -- query vectors for each attention head $1 \leq j \leq k$ (Equations \ref{eq:tr4} and \ref{eq:yappr}). Therefore, $D_2 = Mk + d_{model}$.
\end{itemize}

$F^{(r)}$ and $G^{(r)}$ are functionals parametrized by $\theta$. That is, they take subsets of $\theta$ corresponding to $r$th layer weights (Equations \ref{eq:tr1}-\ref{eq:tr5}). $F^{(r)}$ is responsible for constructing $\mathbf{T}^{(r - 1)}$ and $\mathbf{\Gamma}^{(r - 1)}$ -- representations preceding prefix-sum computation, while $G^{(r)}$ finalizes $\mathrm{MultiHead}\text{-}\mathrm{Att}$ computation (\ref{eq:tr3}) and includes the feed-forward block (\ref{eq:tr5}).

Importantly, $F^{(r)}$ and $G^{(r)}$ are applied \textbf{rowwise}, i.e. (\ref{eq:gtr1}, \ref{eq:gtr3}) can be rewritten as
\begin{gather}
    \forall 1 \leq l \leq L: \mathbf{T}^{(r - 1)}_l, \mathbf{\Gamma}^{(r - 1)}_l = F^{(r)} (\mathbf{X}^{(r - 1)}_l; \theta), \\
    \forall 1 \leq l \leq L: \mathbf{X}^{(r)}_l = G^{(r)} (\mathbf{U}^{(r - 1)}_l, \mathbf{\Gamma}^{(r - 1)}_l ; \theta).
\end{gather}
Hence, \textbf{the only place where the information is propagated across the sequence dimension is the prefix-sum operation (\ref{eq:gtr2})}.

The representation (\ref{eq:gtr1}-\ref{eq:gtr3}) encapsulates architecture details of the Transformer inside $\{ F^{(1)}, G^{(1)}, \dots, F^{(s)}, G^{(s)} \}$. In fact, the representation (\ref{eq:gtr1}-\ref{eq:gtr3}) holds for various possible modifications of the specification (\ref{eq:tr1}-\ref{eq:tr5}), proposed in the literature. This includes, but is not limited by the different positioning of layer normalization \cite{pre-ln,transformer}, adding a stabilizing gating mechanism \citep{stabilizing}, weight sharing across layers \citep{albert} or reversible Transformer layers \citep{reformer}. Therefore, we further analyse the generic, compact notation (\ref{eq:gtr1}-\ref{eq:gtr3}) together with the autoregressive loss formulation (\ref{eq:loss}).

\subsection{Forward Computation} \label{sec:fp}

Suppose the memory budget is not enough to perform a complete forward pass through Performer (Equations \ref{eq:gtr1}-\ref{eq:gtr3} for $r = 1, \dots, s$), because the input sequence length $L$ is too big. We show that instead we can emulate the full forward computation under the memory needed for a forward pass through the input of length $C \leq L$, plus a small addition. $1 \leq C \leq L$ is arbitrary and user-defined.

Split each matrix $\mathbf{X}^{(r)}, \mathbf{T}^{(r)}, \mathbf{\Gamma}^{(r)}, \mathbf{U}^{(r)}$, into $N$ slices of size at most $C$ along the vertical axis ($N = \lceil L / C \rceil$): for each $\forall 1 \leq n \leq N$,
\begin{gather*}
    \mathbf{X}^{(r,n)} = (\mathbf{X}^{(r)}_{A_n + l})_{l = 1}^{B_n} \in \mathbb{R}^{B_n \times d_{model}}, \\
    \mathbf{T}^{(r,n)} = (\mathbf{T}^{(r)}_{A_n + l})_{l = 1}^{B_n}, \mathbf{U}^{(r,n)} = (\mathbf{U}^{(r)}_{A_n + l})_{l = 1}^{B_n} \in \mathbb{R}^{B_n \times D_1}, \\
    \mathbf{\Gamma}^{(r,n)} = (\mathbf{\Gamma}^{(r)}_{A_n + l})_{l = 1}^{B_n} \in \mathbb{R}^{B_n \times D_2},
\end{gather*}
where $A_n = (n - 1) C$ and by $B_n$, $1 \leq n \leq N$, we denote the size of $n$th slice: $B_u = C$ for $u < N$, $B_N \leq C$. Based on (\ref{eq:gtr1}-\ref{eq:gtr3}), we conclude that for each $1 \leq n \leq N$ and $1 \leq r \leq s$ the following recurrence holds:
\begin{gather}
    \mathbf{T}^{(r - 1, n)}, \mathbf{\Gamma}^{(r - 1, n)} = F^{(r)} (\mathbf{X}^{(r, n)}; \theta), \label{eq:cgtr1} \\
    \mathbf{U}^{(r - 1, n)} = \mathbf{1}_{B_n} \!\! \times \! ( \mathbf{U}^{(r - 1, n - 1)}_{B_{n - 1}} )^\top \! + \mathrm{PS} ( \mathbf{T}^{(r - 1, n)} ), \label{eq:cgtr2} \\
    \mathbf{X}^{(r, n)} = G^{(r)} (\mathbf{U}^{(r - 1, n)}, \mathbf{\Gamma}^{(r - 1, n)} ; \theta). \label{eq:cgtr3}
\end{gather}
Here $\mathbf{1}_{B_n} \in \mathbb{R}^{B_n}$ is a vector of $B_n$ ones and we denote $\mathbf{U}^{(r - 1, 0)}_{B_0} = \mathbf{0}_{D_1}$ (a vector of $D_1$ zeros).

Now, instead of iterating over $r = 1, \dots s$ and computing (\ref{eq:gtr1}-\ref{eq:gtr3}) for the whole sequence at once, we \textbf{first iterate over $n = 1, \dots, N$ and then iterate over $r = 1, \dots, s$ in a nested loop} to compute (\ref{eq:cgtr1}-\ref{eq:cgtr3}). As can be deduced from the (\ref{eq:cgtr1}-\ref{eq:cgtr3}), we only need to maintain the current value of $(\mathbf{U}^{(r - 1, n - 1)}_{B_{n - 1}} )_{r = 1}^s \in \mathbb{R}^{s \times D_1}$ in the outer iteration over $n$.

Denote $\mathcal{B}^{(n)} = (\mathbf{U}^{(r - 1, n)}_{B_{n}} )_{r = 1}^s \in \mathbb{R}^{s \times D_1}$, $0 \leq n \leq N$. The memory-efficient algorithm for the forward pass is as follows. First, initialize $\mathcal{L} = 0$ and $\mathcal{B}^{(0)} = \mathbf{0}_{r \times D_1}$. Then, iterate over $n = 1, \dots, N$ and maintain the current value of $\mathcal{B}^{(n - 1)}$. During each iteration, compute $\mathbf{X}^{(0, n)} = (\mathrm{emb} (\mathbf{p}_{A_n + l}, A_n + l))_{l = 1}^{B_n}$. Then iterate over $r = 1, \dots, s$, where compute (\ref{eq:cgtr1}-\ref{eq:cgtr3}) and update $\mathcal{B}^{(n)}_r = \mathbf{U}^{(r - 1, n)}_{B_n}$. Finally, compute $\mathbf{X}^{(out,n)} = \mathbf{X}^{(s,n)} \mathbf{W}^{(out)} + \mathbf{b}^{(out)}$ and update $\mathcal{L} \,+\!\!= \mathcal{L}^{(n)} (\mathbf{X}^{(out,n)})$, where we denote
\begin{equation*}
    \mathcal{L}^{(n)} (\mathbf{X}^{(out,n)}) = (L - 1)^{-1} \sum_{l = 1}^{B_n} \mathcal{L}_{A_n + l} (\mathbf{X}^{(out,n)}_l) .
\end{equation*}

By the end of the iteration over $n$, the correct loss value (\ref{eq:loss}) is computed. As a result, the forward pass takes $O(L)$ serial time or $O((L / C) \log C)$ parallel time and consumes only $O(C)$ memory. This is in addition to the input sequence $\mathbf{p} \in \Sigma^L$ storage, which is $O(L)$ in principle, however the constant is negligibly small. For instance, if $\mathbf{p}$ is a flattened image or an ASCII text string, then it occupies precisely $L$ bytes in memory. The $\log C$ term in the parallel time complexity is due to the parallel prefix-sum algorithm taking logarithmic time, as discussed in Subsection \ref{sec:sa}.

\subsection{Back-Propagation and the Final Algorithm}

\begin{algorithm}[t!]
\caption{Low-memory emulation of the forward-backward pass. See Algorithm \ref{alg:upd} for $\mathrm{updateProc}$. Compared to notation from the text, redundant indices are dropped and tensor names are reused here and in the Algorithm \ref{alg:upd}.}
\label{alg:1}
\begin{algorithmic}
\STATE {\bfseries Input:} $\mathbf{p} \in \Sigma^L$, $\theta \in \mathbb{R}^{n_{param}}, C \in \mathbb{N}$ .
\STATE {\bfseries Output:} loss $\mathcal{L}$, gradient $\nabla_\theta \mathcal{L}$.

\STATE Initialize $\mathcal{L} := 0, \mathcal{B} := \mathbf{0}_{r \times D_1}$;

\algorithmicfor \, $n = 1$ {\bfseries to} $N$ \algorithmicdo \, $\mathrm{updateProc} (n, \mathrm{False})$; \algorithmicend\,\algorithmicfor

\STATE Initialize $\nabla_\theta \mathcal{L} := \mathbf{0}_{n_{param}}, \mathcal{G} := \mathbf{0}_{r \times D_1}$;

\algorithmicfor \, $n = N$ {\bfseries to} $1$ \algorithmicdo \, $\mathrm{updateProc} (n, \mathrm{True})$; \algorithmicend\,\algorithmicfor

\STATE {\bfseries Return} $\mathcal{L}$, $\nabla_\theta \mathcal{L}$ .
\end{algorithmic}
\end{algorithm}

\begin{algorithm}[t!]
\caption{$\mathrm{updateProc}$ procedure.}
\label{alg:upd}
\begin{algorithmic}
\STATE {\bfseries Input:} $n \in \mathbb{N}$, binary flag $\mathrm{onBackprop}$ .

\algorithmicif \, $\mathrm{onBackprop}$ \algorithmicthen\,Initialize $\Phi := 0$; \algorithmicend\,\algorithmicif

\STATE $\mathbf{X} := (\mathrm{emb} (\mathbf{p}_{A_n + l}, A_n + l))_{l = 1}^{B_n}$;

\FOR {$r = 1$ {\bfseries to} $s$}
    \STATE Compute $\mathbf{T}, \mathbf{\Gamma} := F^{(r)} (\mathbf{X}; \theta)$;

    \algorithmicif \, $\mathrm{onBackprop}$ \algorithmicthen\,Update $\mathcal{B}_r -\!\!= \sum_{l = 1}^{B_n} \mathbf{T}_l$; \algorithmicend\,\algorithmicif

    \STATE Set $\mathbf{U} := \mathbf{1}_{B_n} \mathcal{B}_r^\top + \mathrm{PS} ( \mathbf{T} )$, $\mathbf{X} := G^{(r)} (\mathbf{U}, \mathbf{\Gamma} ; \theta)$;

    \IF {$\mathrm{onBackprop}$}
        \STATE Update $\Phi +\!\!= \mathcal{G}_r^\top \mathbf{U}_{B_n}$;
    \ELSE
        \STATE Update $\mathcal{B}_r := \mathbf{U}_{B_n}$;
    \ENDIF

\ENDFOR

\STATE Set $\mathcal{L}^{(upd)} := \mathcal{L}^{(n)} (\mathbf{X} \mathbf{W}^{(out)} + \mathbf{b}^{(out)} )$;

\IF {$\mathrm{onBackprop}$}
    \STATE Update $\Phi +\!\!= \mathcal{L}^{(upd)}$;

    \STATE Compute $\nabla_\theta \Phi, \nabla_{\mathcal{B}} \Phi$ through auto-differentiation;

    \STATE Update $\nabla_\theta \mathcal{L} +\!\!= \nabla_\theta \Phi$, \,\, $\mathcal{G} := \nabla_{\mathcal{B}} \Phi$;
\ELSE
    \STATE Set $\mathcal{L} +\!\!= \mathcal{L}^{(upd)}$;
\ENDIF
\end{algorithmic}
\end{algorithm}

The goal of a backward pass is to compute gradient $\nabla_\theta \mathcal{L}$ of the loss function with respect to parameters $\theta$. One can just perform automatic differentiation \citep{autograd} (implemented in Tensorflow \citep{tensorflow} and Pytorch \citep{pytorch}) through the computation graph induced by the memory-efficient forward pass algorithm from Subsection \ref{sec:fp}. However, such backward pass would need to store all intermediate tensors produced during the forward pass,  resulting in $O(L)$ memory complexity as a function of $L$ and $C$. Instead, we propose a back-propagation algorithm which has the same time and memory complexity as the efficient forward pass.

Let $\theta^{(1)} = \dots = \theta^{(N)} = \theta$ be results of a symbolic ``identity operation'' performed on $\theta$, so that for all $1 \leq n \leq N$, $\theta^{(n)}$ is used instead of $\theta$ in (\ref{eq:cgtr1}-\ref{eq:cgtr3}). Then the total gradient of $\theta$ has the form $\nabla_\theta \mathcal{L} = \nabla_{\theta^{(1)}} \mathcal{L} + \dots + \nabla_{\theta^{(N)}} \mathcal{L}$. In Appendix \ref{sec:deriv} we derive an expression for $\nabla_{\theta^{(n)}} \mathcal{L}$, $1 \leq n \leq N$. Namely, denote $\mathcal{G}^{(n)} = \nabla_{\mathcal{B}^{(n)}} \mathcal{L}$, then $\nabla_{\theta^{(n)}} \mathcal{L} = \nabla_{\theta^{(n)}} \Phi^{(n)} ( \theta^{(n)}, \mathcal{B}^{(n - 1)}, \mathcal{G}^{(n)} )$, where $\Phi^{(n)}: \mathbb{R}^{n_{param}} \times \mathbb{R}^{s \times D_1} \times \mathbb{R}^{s \times D_1} \to \mathbb{R}$,
\begin{equation*}
   \Phi^{(n)} ( \theta^{(n)}, \mathcal{B}^{(n - 1)}, \mathbf{Z} ) = \mathcal{L}^{(n)} (\mathbf{X}^{(out,n)}) + \sum_{r = 1}^s \mathbf{Z}_r^\top \mathcal{B}^{(n)}_r .
\end{equation*}
In $\Phi^{(n)}$'s definition, $\mathbf{X}^{(out,n)} = \mathbf{X}^{(s,n)} \mathbf{W}^{(out)} + \mathbf{b}^{(out)}$ and $\mathcal{B}^{(n)} = (\mathbf{U}^{(r - 1, n)}_{B_n} )_{r = 1}^s$ are results of (\ref{eq:cgtr1}-\ref{eq:cgtr3}) iteration over $r = 1, \dots, s$ with parameters $\theta = \theta^{(n)}$ and $(\mathbf{U}^{(r - 1, n - 1)}_{B_{n - 1}} )_{r = 1}^s$ equal to $\Phi^{(n)}$'s second argument $\mathcal{B}^{(n - 1)}$. Gradient $\nabla_{\theta^{(n)}} \Phi^{(n)}$ can be computed by automatic differentiation through the computation graph induced by $\Phi^{(n)}$.

An efficient way to compute and sum up all $\nabla_{\theta^{(n)}} \mathcal{L}$ is to iterate in a backward direction $n = N, \dots, 1$ and to maintain current values of $\mathcal{B}^{(n)}, \mathcal{G}^{(n)}$. $\mathcal{B}^{(N)}$ is known after the end of the forward pass, and for each $1 \leq n \leq N$,
\begin{equation}
    \mathcal{B}^{(n - 1)} = \mathcal{B}^{(n)} - \sum_{l = 1}^{B_n} ( \mathbf{T}^{(r - 1,n)}_l )_{r = 1}^s . \label{eq:recomp}
\end{equation}
Further, in Appendix \ref{sec:deriv} we show that $\mathcal{G}^{(N)} = \mathbf{0}_{r \times D_1}$ and, for each $1 \leq n \leq N$,
\begin{equation}
    \mathcal{G}^{(n - 1)} = \nabla_{\mathcal{B}^{(n - 1)}} \Phi^{(n)} (\theta^{(n)}, \mathcal{B}^{(n - 1)}, \mathcal{G}^{(n)}). \label{eq:gradupd}
\end{equation}
By a single auto-differentation through $\Phi^{(n)}$ we can compute $\nabla_{\theta^{(n)}} \mathcal{L} = \nabla_{\theta^{(n)}} \Phi^{(n)}$ and the update (\ref{eq:gradupd}).

Observe that, if $\mathbf{w}$ is some vector of length $B_n$ and $h$ is some scalar function of $\mathbf{v} = \mathrm{PS} (\mathbf{w})$, then for all $1 \leq l \leq B_n$ : $\nabla_h \mathbf{w}_l = \sum_{l' = t}^{B_n} \nabla_h \mathbf{v}_{l'}$. In other words, the gradient through $\mathrm{PS} (\cdot)$ is another prefix sum computed backwards. Hence, auto-differentiation through $\Phi^{(n)}$ takes the same parallel time $O( \log C )$, serial time $O(L)$ and memory $O(C)$, as the forward computation of $\Phi^{(n)}$. Since during the whole back-propagation algorithm, we only store and update tensors $\mathcal{B}^{(n)}, \mathcal{G}^{(n)}$, whose size doesn't depend on $L$ and $C$, this results in total $O((L / C) \log C)$ parallel time, $O (L)$ serial time and $O(C)$ memory in addition to $\mathbf{p}$ storage. A full description of the forward-backward pass is presented in Algorithm \ref{alg:1}. Figure \ref{fig:1}b is an illustration of the algorithm.

\subsection{Analysis of the Running Time and Memory} \label{sec:tradeoff}

As we have shown, Performer can be trained in parallel time $O((L / C) \log C)$ and $O(C)$ memory in addition to the input sequence $\mathbf{p}$ storage. Hence, $C$ is a tradeoff parameter: when $C$ is maximal ($C = L$), the model is fully-parallelized along the sequence dimension, therefore resulting in the fastest execution. Whereas minimal $C = 1$ corresponds to step-by-step processing, i.e. a fully-sequential regime which doesn't benefit from parallelized computations on GPU or TPU, but consumes $O(1)$ memory as a function of $L$.

It can be seen that during the forward pass, Algorithm \ref{alg:1} requires as many total FLOPs as the naive forward pass through (\ref{eq:cgtr1}-\ref{eq:cgtr3}). As for the backward pass, for each $1 \leq n \leq N$, the forward pass through $n$'s slice is repeated for symbolical construction of $\Phi^{(n)}$ (see Algorithm \ref{alg:upd}), and then back-propagation is run through $\Phi^{(n)}$. In addition, a backward update of $\mathcal{B}^{(n)}$ (\ref{eq:recomp}) is computed, taking precisely $B_n s M (d + 1) k$ ``add'' operations. Hence, we conclude that \textbf{Algorithm \ref{alg:1} requires as many FLOPs as two forward and one backward pass through (\ref{eq:cgtr1}-\ref{eq:cgtr3}) for the whole sequence $\mathbf{p}$} plus $L s M (d + 1) k = L s M d_{model} + L s M k$ FLOPs. To characterize this addition, assuming that typically $d_{ff} = 4 d_{model}$ in practice, observe that applying linear operators in (\ref{eq:tr1}-\ref{eq:tr5}) alone requires
\begin{equation*}
    3 L s d^2_{model} + 2 L s d_{model} d_{ff} = 11 L s d_{model}^2
\end{equation*}
FLOPs. This is much bigger than $L s M d_{model} + L s M k$, since $M$ is much smaller than $d_{model}$ in practice \citep{performer,transfrnn}.

Since the back-propagation takes roughly $5$ times more FLOPs than the forward pass \citep{autograd}, we conclude that \textbf{memory efficiency of Algorithm \ref{alg:1} results in a small constant-time increase in FLOPs}. The FLOPs count has a direct effect on  energy consumption \citep{litetransf}, a crucial factor for on-device applications.

Further analysis of  Algorithm \ref{alg:1} reveals that the $C = 1$ regime requires \textbf{as much memory as if Transformer were applied to a sequence of length 1} plus exactly $2 s d_{model} (M + 1)$ floats for storing $\mathcal{B}, \mathcal{G}$. For comparison, the subset of $\theta$ corresponding to matrix parameters in self-attention and feed-forward blocks (\ref{eq:tr1}-\ref{eq:tr5}), occupies
\begin{equation*}
    3 s d_{model}^2 + 2 s d_{model} d_{ff} = 11 s d_{model}^2
\end{equation*}
floats. Again, this is much bigger than $2 s d_{model} (M + 1)$, since $M$ is much smaller than $d_{model}$ in practice.

To understand these fruitful properties, we perform a conceptual comparison of Performer, recurrent neural networks (RNNs, \citet{lstm,gru}) and residual architectures (e.g. Neural ODEs, \citet{neuralode}), which are also used for sequence processing. The $r$th layer of all models has the following form for $1 \leq l \leq L$:
\begin{align}
    \text{RNN}:& \quad \mathbf{X}^{(r)}_l = f^{(r)}(\textcolor{red}{\mathbf{X}^{(r)}_{l - 1}}, \mathbf{X}^{(r - 1)}_l), \label{eq:comp1} \\
    \text{Residual}:& \quad \mathbf{X}^{(r)}_l = \mathbf{X}^{(r)}_{l - 1} + f^{(r)}(\textcolor{red}{\mathbf{X}^{(r)}_{l - 1}}, \mathbf{X}^{(r - 1)}_l) , \\
    \text{Performer}:& \quad \mathbf{X}^{(r)}_l = \mathbf{X}^{(r)}_{l - 1} + f^{(r)}(\mathbf{X}^{(r - 1)}_l) . \label{eq:comp3}
\end{align}
Here $f^{(r)}$ is some nonlinear map. Observe that Performer is the only architecture where $\mathbf{X}^{(r)}_l$ depends linearly on $\mathbf{X}^{(r)}_{l - 1}$. It's not hard to see that Algorithm \ref{alg:1} can be applied to any architecture of type (\ref{eq:comp3}). Despite the update's simplicity, Performer appears to work very well in challenging real-life setups, and, as shown by \citet{performer}, can approximate any conventional Transformer with exponential self-attention. See Table \ref{tab:1} for a complexity comparison of all discussed architectures and the proposed algorithm.

\begin{table}[h!]
\caption{Complexity for the exact forward-backward pass as functions of sequence length $L$ and the tradeoff parameter $C \leq L$ (for Performer). The indicated memory complexity is in addition to the input sequence $\mathbf{p}$ storage. The serial time complexity for Performer is reported for the version with iterative $\mathrm{PS}(\cdot)$ computation (as in \citealp{transfrnn}), while the parallel time is reported for the parallel prefix sum (as in \citealp{performer}). For both methods, memory complexity is the same, though the constant is smaller for the iterative version.}
\label{tab:1}
\vskip 0.15in
\begin{center}
\begin{small}
\begin{sc}
\begin{tabular}{lccr}
\toprule
Model & \makecell{Serial \\ time} & \makecell{Parallel \\ time} & Memory \\
\midrule
RNN & $O(L)$ & $O(L)$ & $O(L)$ \\
Residual NN & $O(L)$ & $O(L)$ & $O(L)$ \\
$\mathrm{Att}^{exp}$ Transf. & $O(L^2)$ & $O(\log L)$ & $O(L^2)$ \\
Performer & $O(L)$ & $O(\log L)$ & $O(L)$ \\
Our algorithm & $O(L)$ & $O(\frac{L}{C} \log C)$ & $O(C)$ \\
Our alg., $C = 1$ & $O(L)$ & $O(L)$ & $\boldsymbol{O(1)}$ \\
\bottomrule
\end{tabular}
\end{sc}
\end{small}
\end{center}
\vskip -0.1in
\end{table}

\begin{figure*}[ht!]
    \centering
    \includegraphics[width=\textwidth]{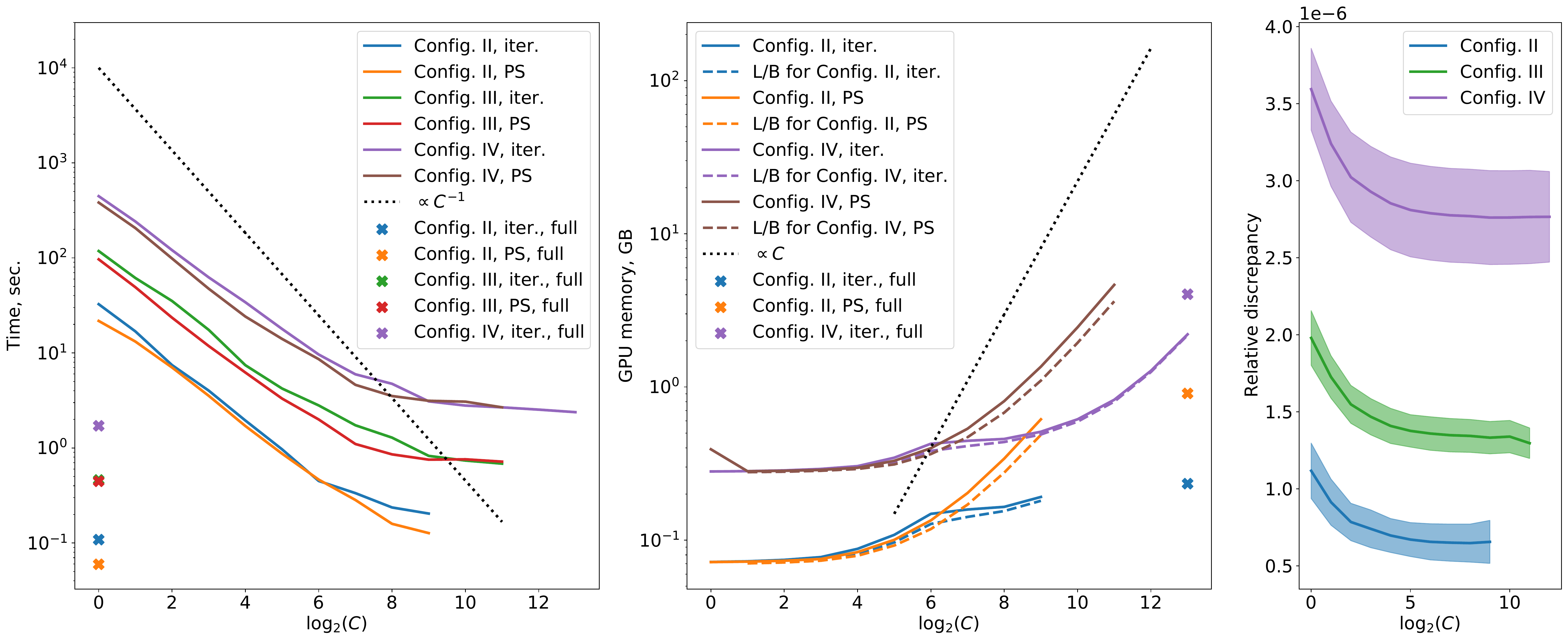}
    \caption{Benchmarks of Algorithm \ref{alg:1}. All plots are averaged over $10$ seeds. ``iter.'' stands for iterative computation of (\ref{eq:pspart}-\ref{eq:yappr}), while ``PS'' is for explicit prefix sum computation in (\ref{eq:pspart}). We don't report time and memory for big values of $C$ in ``Config. IV, PS'' setup and for ``Config. IV, full'' setup, because these runs resulted in memory overflow. \textbf{(Left)} Time dependence on $C$. Crosses indicate horizontal time levels for corresponding full memory-inefficient methods. The dotted line indicates $\propto C^{-1}$ tangent in logarithmic scale. \textbf{(Middle)} Memory dependence on $C$. Again, crosses are for horizontal levels of full-sequence methods and the dotted line indicates $\propto C$ tangent. We do not report curves for config. III, because they completely match curves for config. IV, which is natural, since $d_{model}$ is the same for both configurations. ``L/B'' stands for a memory lower bound computed by processing input of length $C$. \textbf{(Right)} Relative gradient discrepancy as a function of $C$, also reporting standard errors.}
    \label{fig:tm}
\end{figure*}

\section{Experiments} \label{sec:exp}

Our main contribution is a new low-memory gradient computation algorithm for the existing Performer architecture. Performers have %, which shows 
very competitive performance among other methods for long sequence modelling \citep{performer,transfrnn,lra}. Hence, in the experimental section, we aim to answer the following questions about using this algorithm in practice:
\begin{enumerate}
    \item Does the theoretical time-memory tradeoff, controlled by $C$, agree with empirical benchmarks of time and memory for $C$ variation?
    \item In precise arithmetic, different values of $C$ lead to the same correct gradient $\nabla_\theta \mathcal{L}$. Does this hold in practice, when finite-precision arithmetic is employed?
    \item Can a model, pre-trained with a bigger value of $C$ (e.g. on a server), be fine-tuned with a smaller $C$ (e.g. on a smartphone)? Does the parameter $C$ affect the performance of training from scratch?
\end{enumerate}

We address each question in detail in the subsections below. In our experiments, we analyse 4 model configurations $(L, d_{model})$: $I = (512, 256)$, $II = (1024, 512)$, $III = (4096, 1024)$, $IV = (16384, 1024)$. In all configurations, we set $d_{ff} = 4 d_{model}$, $k = d_{model}/64$ (number of heads), $s = 3$ (number of layers). We set $M = d$ and employ $g (\mathbf{x}) = (\mathbf{x}_i^2)_{i = 1}^d$ elementwise-quadratic feature mapping in (\ref{eq:mcatt}), which we find to work well in practice. In all experiments $\Sigma = \{ 0, \dots, 255 \}$ and batch size is set to $1$, i.e. we analyse a setup where gradient accumulation cannot be used to decrease memory, and therefore our algorithm is crucial. Our code is in PyTorch 1.7. To ensure that reproduction of experiments is accessible for a wider audience, we use a single NVIDIA Tesla P100 GPU with 16 GB memory for each experiment.

\begin{figure*}[h!]
    \centering
    \includegraphics[width=\textwidth]{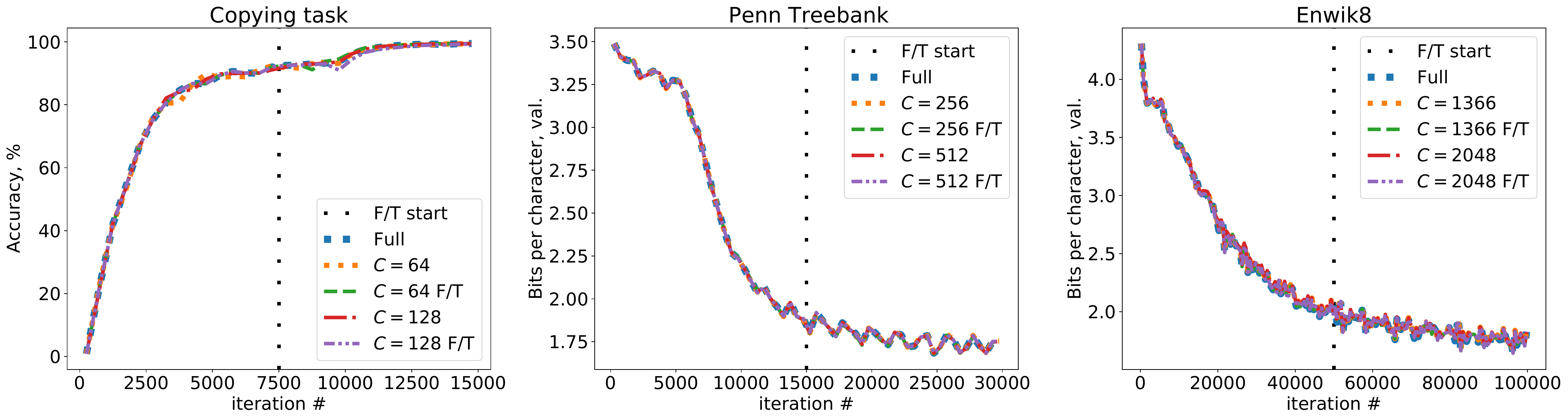}
    \caption{Learning curves for three language modelling setups. We report accuracy on a newly generated data samples for Copying task, and bits-per-character metric on validation examples for Penn Treebank and Enwik8. F/T stands for ``fine-tuning''. \textbf{All curves are almost indistinguishable, confirming correctness and backward-compatibility} of gradients computed via memory-efficient Algorithm \ref{alg:1}.}
    \label{fig:tr}
\end{figure*}

\subsection{Empirical Benchmarking of the Tradeoff}

We run Algorithm \ref{alg:1} for configurations II-IV and different powers of 2 as $C$. We use input strings sampled randomly from $\Sigma^L$. In order to characterize the time-memory tradeoff, we measure wall-clock time and peak GPU memory for a single gradient evaluation. We use the \texttt{torch.cuda.max\_memory\_allocated} function to report peak GPU memory.

As discussed in Section \ref{sec:sa}, there are two methods to compute (\ref{eq:pspart}-\ref{eq:yappr}): the first (iterative) method doesn't compute and store tensors (\ref{eq:yappr}) explicitly, resulting in smaller memory consumption at a cost of less parallelization, while the second one computes tensors (\ref{eq:yappr}) using the parallel prefix sum algorithm, therefore operating faster, but using more memory. The same methods can be applied for the memory-efficient algorithm when computing (\ref{eq:cgtr2}-\ref{eq:cgtr3}) updates. We implement and benchmark both methods as part of the algorithm. For the explicit prefix-sum method, we find that the \texttt{torch.cumsum} function works faster and consumes less memory than our custom implementation of the parallel prefix sum algorithm. We attribute this to hardware-optimized low-level implementation of the native function, and use this function in experiments. As for the iterative algorithm, we implement its ``block'' version, when, instead of iterating $l$ one-by-one, we iterate through blocks of small size (see details in Appendix \ref{sec:biter}). This way, the algorithm has a smaller constant in $O(L)$ time complexity and bigger constant in a ``small'' $O(d M)$ term of the memory complexity (assuming that $d, M \ll L$).

For a fixed value of $C$, in addition to benchmarking memory of Algorithm \ref{alg:1}, we also report memory of the naive gradient computation run on a string of length $C$, sampled uniformly from $\Sigma^C$. This is to confirm that memory consumption of Algorithm \ref{alg:1} is just slightly above the full computation on the input of length $C$.

Results are reported in Figure \ref{fig:tm} (left, middle). We observe significant improvements in memory consumption compared to the full computation, as $C$ decreases. As $C$ converges to $2^0 = 1$, the remaining memory consumption can be attributed to storage of the model's parameters $\theta$. Time follows two regimes: declining fast as $C$ grows (meaning that prefix sums are parallelized) and declining slower for big values of $C$ (meaning that the practical limit of parallelization is reached). Memory scales slower than $O(C)$, as $C$ increases. We attribute this effect to details of PyTorch internal implementation. Interestingly, we find that iterative version of (\ref{eq:pspart}-\ref{eq:yappr}) computation works only slightly slower than prefix-sum version, while consuming much less memory. Finally, Algorithm \ref{alg:1} consumes slightly more memory in practice than the full method run on the input of length $C$.

\subsection{Effects of Finite-Precision Arithmetic}

Since the iterative version of (\ref{eq:pspart}-\ref{eq:yappr}) computation results in a good balance between time and memory of Algorithm \ref{alg:1}, we use it in our subsequent experiments. To quantify finite-precision effects, we plot \textit{relative discrepancy} $\| \nabla_{\theta}^{(C)} \mathcal{L} - \nabla_{\theta}^{(full)} \mathcal{L} \|_2 / \| \nabla_{\theta}^{(full)} \mathcal{L} \|_2$ between the gradient $\nabla_{\theta}^{(C)}$ produced by Algorithm \ref{alg:1}, and the gradient $\nabla_{\theta}^{(full)} \mathcal{L}$ produced by full-input computation. Figure \ref{fig:tm} illustrates results for randomly initialized models. We observe a very small discrepancy (of order $10^{-6}$--$10^{-5}$), confirming the correctness of Algorithm \ref{alg:1}. The discrepancy is slightly increasing as $C$ decreases, which can be attributed to effects of finite-precision arithmetic.

\subsection{Training from Scratch and Fine-tuning}

To confirm backward compatibility of Algorithm \ref{alg:1} during training, we consider three language modelling setups: Copying task, symbol-level Penn Treebank and Enwik8.

For the Copying task, we follow the setup from \citep{reformer,transfrnn}, sampling inputs as $0\omega0\omega$, where $\omega$ is drawn uniformly from $(\Sigma \setminus \{ 0 \})^{L / 2 - 1}$. In this setup, we only aggregate cross-entropy loss from the second half of the input, so the task is to reproduce the first half. We include the Copying task as an example setup where long-range information propagation is crucial, and the heuristic of ``chunking'' the input into smaller segments would fail to solve the task.

We use model configurations I, II, III for the Copying task, Penn Treebank and Enwik8 respectively, resulting in sequence lengths $L = 512, 1024, 4096$ respectively. For each setup, we compare training with full gradient computation, and training equipped with memory-efficient gradient computation via Algorithm \ref{alg:1} using various values of $C$. In addition, we consider a ``fine-tuning'' regime, when the first half of iterations is run using the full algorithm, and the second half is run using Algorithm \ref{alg:1}. Figure \ref{fig:tr} demonstrates results: all methods result in the same, indistinguishable performance. \textbf{This confirms that memory-efficient gradient computation can be used both for training from scratch, and for fine-tuning, e.g. on a low-memory device}. Table \ref{tab:2} quantifies the memory savings and time tradeoff in all setups.  Additional experimental details and results (bigger version of Figure \ref{fig:tr}, bits-per-character for the Copying task and train set performance for Penn treebank and Enwik8) can be found in Appendix \ref{sec:addexp}.

\section{Related Work and Extensions}

\textbf{Compatibility with other memory-optimization techniques.} Observe that the specification (\ref{eq:gtr1}-\ref{eq:gtr3}) is compatible with the reversible layer design from \citep{reformer}, when the sparse self-attention is replaced with the linear self-attention\footnote{See e.g. \texttt{CausalFavor} class in \href{https://github.com/google/trax/blob/master/trax/layers/research/sparsity.py}{https://github.com/ google/trax/blob/master/trax/layers/research/sparsity.py}, which is compatible with the official Reformer code.}. This can bring more memory savings, since one doesn't need to store the whole symbolic $\Phi^{(n)}$ during the backward pass. Checkpointing techniques \citep{logckpt,gckpt} can also be used to reduce the memory consumption for storing $\Phi^{(n)}$'s graph, though at the cost of a longer execution time. The gradient accumulation technique \citep{gradacc} is also compatible with Algorithm \ref{alg:1}, i.e. one can combine both methods to ``collapse'' batch and sequence dimensions simultaneously. Moreover, our algorithm is compatible with distillation \citep{distil}, since it can be run on a distilled model.

\textbf{Comparison with \citep{transfrnn}.} \citet{transfrnn} mention that a single self-attention block can be evaluated in $O(1)$ additional memory. However, one still needs to store $L$ intermediate states, e.g. in the feedforward block. Hence, the full memory complexity is still $O(L)$. In contrast, our method optimizes memory consumption along the sequence dimension for the whole multilayer model.

\textbf{Extension to Transformers with dropout.} Dropout \citep{dropout} is a popular regularization technique. It is used with Transformers when the train dataset is small enough to cause overfitting (e.g. it wasn't used with GPT-2, trained on a massive dataset). Our algorithm can be extended to stochastic computation graphs with dropout. For that, use separate random seeds to generate dropout masks for each slice $1 \leq n \leq N$, and reuse these seeds two times during the forward and backward pass through the $n$th slice.

\section{Conclusion}

We proposed an algorithm for memory-efficient back-propagation through a Performer. The algorithm reduces memory consumption along the sequence dimension, and can, therefore, be used for long-sequence training. The algorithm: (1) is completely backward-compatible, since it computes precise gradients and does not involve approximation, (2) does not require many additional computations, and (3) enables user control over the tradeoff between time and memory consumption.

\section{Acknowledgments}

We thank Tom Weingarten and Tamas Sarlos for many fruitful discussions.

Valerii Likhosherstov acknowledges support from the Cambridge Trust and DeepMind. Adrian Weller acknowledges support from The Alan Turing Institute under
EPSRC grant EP/N510129/1 and U/B/000074, and the Leverhulme Trust via CFI.

\begin{table}
\caption{Time per iteration (averaged over 1000 iterations) and peak GPU memory. CT -- Copying task, PTB -- Penn Treebank.}
\label{tab:2}
\vskip 0.15in
\begin{center}
\begin{small}
\begin{sc}
\begin{tabular}{ccc}
\toprule
Setup, $L$, $C$ & \makecell{Time per \\ iter. (sec.)} & \makecell{GPU me- \\ mory (GB)} \\
\midrule
CT 512, full & 0.0474 & 0.0449 \\
CT 512, 128 & 0.0921 & 0.0425 \\
CT 512, 64 & 0.1228 & \textbf{0.0374} \\
\midrule
PTB 1024, full & 0.1377 & 0.300 \\
PTB 1024, 512 & 0.2526 & 0.257 \\
PTB 1024, 256 & 0.3060 & \textbf{0.231} \\
\midrule
Enwik8 4096, full & 0.4598 & 1.513 \\
Enwik8 4096, 2048 & 0.7922 & 1.085 \\
Enwik8 4096, 1366 & 0.8654 & \textbf{0.909} \\
\bottomrule
\end{tabular}
\end{sc}
\end{small}
\end{center}
\vskip -0.1in
\end{table}

\bibliography{example_paper}
\bibliographystyle{icml2020}

\newpage
\onecolumn
\appendix

\section{Derivation of gradient expressions} \label{sec:deriv}

$\theta^{(n)}$ doesn't affect terms $\mathbf{\mathcal{L}}^{(1)} (\mathbf{X}^{(out, 1)}), \dots, \mathbf{\mathcal{L}}^{(n - 1)} (\mathbf{X}^{(out, n)})$, so corresponding gradients are zero:
\begin{equation*}
    \nabla_{\theta^{(n)}} \mathcal{L} = \nabla_{\theta^{(n)}} \sum_{n' = n}^N  \mathcal{L}^{(n')} (\mathbf{X}^{(out,n')}).
\end{equation*}
Similarly, $\mathcal{B}^{(n)}$ does not affect $\mathbf{\mathcal{L}}^{(1)}, \dots, \mathbf{\mathcal{L}}^{(n)}$, so
\begin{equation*}
    \mathcal{G}^{(n)} = \nabla_{\mathcal{B}^{(n)}} \mathcal{L} = \nabla_{\mathcal{B}^{(n)}} \sum_{n' = n + 1}^N  \mathcal{L}^{(n)} (\mathbf{X}^{(out,n')}).
\end{equation*}
In particular,
\begin{equation*}
    \mathcal{G}^{(N)} = \nabla_{\mathcal{B}^{(N)}} \mathcal{L} = \mathbf{0}_{r \times D_1}.
\end{equation*}

For all $1 \leq n < n' \leq N$, $\theta^{(n)}$ and $\mathcal{B}^{(n - 1)}$ affect $\mathcal{L}^{(n')}$ only through $\mathcal{B}^{(n)}$, so according to the chain rule
\begin{gather*}
    \nabla_{\theta^{(n)}} \sum_{n' = n + 1}^N \mathcal{L}^{(n')} (\mathcal{X}^{(out,n')}) = \sum_{r = 1}^s \frac{\partial \mathcal{B}^{(n)}_r}{\partial \theta^{(n)}}^\top \times \nabla_{\mathcal{B}^{(n)}_r} \sum_{n' = n + 1}^N \mathcal{L}^{(n')} (\mathcal{X}^{(out,n')})
    = \sum_{r = 1}^s \frac{\partial \mathcal{B}^{(n)}_r}{\partial \theta^{(n)}}^\top \times \nabla_{\mathcal{B}^{(n)}_r} \mathcal{L} , \\
    %%%%
    %%%%
    \forall 1 \leq r' \leq s : \nabla_{\mathcal{B}^{(n - 1)}_{r'}} \sum_{n' = n + 1}^N \mathcal{L}^{(n')} (\mathcal{X}^{(out,n')}) = \sum_{r = 1}^s \frac{\partial \mathcal{B}^{(n)}_r}{\partial \mathcal{B}^{(n - 1)}_{r'}}^\top \times \nabla_{\mathcal{B}^{(n)}_r} \sum_{n' = n + 1}^N \mathcal{L}^{(n')} (\mathcal{X}^{(out,n')}) \\
    = \sum_{r = 1}^s \frac{\partial \mathcal{B}^{(n)}_r}{\partial \mathcal{B}^{(n - 1)}_{r'}}^\top \times \nabla_{\mathcal{B}^{(n)}_r} \mathcal{L} ,
\end{gather*}
where $\frac{\partial \square}{\partial \square}$ denotes Jacobian matrices. Further, for all $1 \leq r \leq s$:
\begin{gather*}
    \frac{\partial \mathcal{B}^{(n)}_r}{\partial \square}^\top \times \nabla_{\mathcal{B}^{(n)}_r} \mathcal{L} = \nabla_\square \biggl( [\mathcal{B}^{(n)}_r]^\top \langle\langle \nabla_{\mathcal{B}^{(n)}_r} \mathcal{L} \rangle\rangle \biggr),
\end{gather*}
where $\square \in \{ \theta^{(n)} \} \cup \{ \mathcal{B}^{(n - 1)}_{r'} \}_{1 \leq r' \leq s}$. $\langle\langle\cdot\rangle\rangle$ denotes a \textit{stop-gradient} operator, i.e. gradients are not propagated inside brackets and the argument is considered as constant.

We conclude that
\begin{gather*}
    \nabla_{\theta^{(n)}} \mathcal{L} = \nabla_{\theta^{(n)}} \mathcal{L}^{(n)} (\mathcal{X}^{(out,n)}) + \nabla_{\theta^{(n)}} \sum_{n' = n + 1}^N \mathcal{L}^{(n')} (\mathcal{X}^{(out,n')}) = \nabla_{\theta^{(n)}} \mathcal{L}^{(n)} (\mathcal{X}^{(out,n)}) + \sum_{r = 1}^s \frac{\partial \mathcal{B}^{(n)}_r}{\partial \theta^{(n)}}^\top \times \nabla_{\mathcal{B}^{(n)}_r} \mathcal{L} \\
    = \nabla_{\theta^{(n)}} \biggl( \mathcal{L}^{(n)} (\mathcal{X}^{(out,n)}) + \sum_{r = 1}^s [\mathcal{B}^{(n)}_r]^\top \langle\langle \nabla_{\mathcal{B}^{(n)}_r} \mathcal{L} \rangle\rangle \biggr) = \nabla_{\theta^{(n)}} \Phi^{(n)} (\theta^{(u)}, \mathcal{B}^{(n - 1)}, \nabla_{\mathcal{B}^{(n)}} \mathcal{L}) \\
    = \nabla_{\theta^{(n)}} \Phi^{(n)} (\theta^{(u)}, \mathcal{B}^{(n - 1)}, \mathcal{G}^{(n)}), \\
    %%%%
    %%%%
    \forall 1 \leq r' \leq s : \mathcal{G}^{(n - 1)}_{r'} = \nabla_{\mathcal{B}^{(n - 1)}_{r'}} \mathcal{L} = \nabla_{\mathcal{B}^{(n - 1)}_{r'}} \mathcal{L}^{(n)} (\mathcal{X}^{(out,n)}) + \nabla_{\mathcal{B}^{(n - 1)}_{r'}} \sum_{n' = n + 1}^N \mathcal{L}^{(n')} (\mathcal{X}^{(out,n')}) \\
    = \nabla_{\mathcal{B}^{(n - 1)}_{r'}} \mathcal{L}^{(n)} (\mathcal{X}^{(out,n)}) + \sum_{r = 1}^s \frac{\partial \mathcal{B}^{(n)}_r}{\partial \mathcal{B}^{(n - 1)}_{r'}}^\top \times \nabla_{\mathcal{B}^{(n)}_r} \mathcal{L} \\
    = \nabla_{\mathcal{B}^{(n - 1)}_{r'}} \biggl( \mathcal{L}^{(n)} (\mathcal{X}^{(out,n)}) + \sum_{r = 1}^s \nabla_\square [\mathcal{B}^{(n)}_r]^\top \langle\langle \nabla_{\mathcal{B}^{(n)}_r} \mathcal{L} \rangle\rangle \biggr) = \nabla_{\mathcal{B}^{(n - 1)}_{r'}} \Phi^{(n)} (\theta^{(n)}, \mathcal{B}^{(n - 1)}, \nabla_{\mathcal{B}^{(n)}} \mathcal{L}) \\
    = \nabla_{\mathcal{B}^{(n - 1)}_{r'}} \Phi^{(n)} (\theta^{(n)}, \mathcal{B}^{(n - 1)}, \mathcal{G}^{(n)}) ,
\end{gather*}
where the second chain of equalities is equivalent to (\ref{eq:gradupd}).

\section{Efficient ``Block'' Computation of (\ref{eq:pspart}-\ref{eq:yappr})} \label{sec:biter}

Denote $\widetilde{\mathbf{Q}} = (g(\mathbf{Q}_l))_{l = 1}^L$, $\widetilde{\mathbf{K}} = (g(\mathbf{K}_l))_{l = 1}^L$, $\mathbf{N} = (\mathbf{R}_l \times \widetilde{\mathbf{Q}}_l)_{l = 1}^L$, $\mathbf{D} = (\mathbf{S}_l^\top \widetilde{\mathbf{Q}}_l)_{l = 1}^L$. \citet{transfrnn} propose the following algorithm for computation of (\ref{eq:pspart}-\ref{eq:yappr}). Initialize buffers $\mathrm{cur}\mathbf{R} = \mathbf{0}_{d \times M}, \mathrm{cur}\mathbf{S} = \mathbf{0}_M$, iterate over $l = 1, \dots, L$ and compute
\begin{align*}
    &\mathrm{cur}\mathbf{R} := \mathrm{cur}\mathbf{R} + \mathbf{V}_l \times \widetilde{\mathbf{K}}_l^\top; \\
    &\mathrm{cur}\mathbf{S} := \mathrm{cur}\mathbf{S} + \widetilde{\mathbf{K}}_l; \\
    &\mathbf{N}_l := \mathrm{cur}\mathbf{R} \times \widetilde{\mathbf{Q}}_l; \\
    &\mathbf{D}_l := \mathrm{cur}\mathbf{S}^\top \times \widetilde{\mathbf{Q}}_l; \\
    &\mathbf{Y}_l := \mathbf{N}_l / \mathbf{D}_l.
\end{align*}
This way, 3d tensor $\mathbf{R} \in \mathbb{R}^{L \times d \times M}$ is not stored in memory explicitly, resulting in $O(L)$ time and $O(L (d + M) + d M)$ memory complexity. In order to have the same memory consumption during back-propagation, \citet{transfrnn} propose the following routine. Keep buffers $\mathrm{cur}\mathbf{R}, \mathrm{cur}\mathbf{S}$ as the result of forward pass, and initialize gradient buffers $\mathrm{grad}\mathbf{R} = \mathbf{0}_{d \times M}, \mathrm{grad}\mathbf{S} = \mathbf{0}_M$. Assuming that $\nabla_\mathbf{N} \mathcal{L} \in \mathbb{R}^{L \times d}, \nabla_\mathbf{D} \mathcal{L} \in \mathbb{R}^L$ are computed using automatic differentiation, iterate in a backward direction $l = L, \dots, 1$ and compute
\begin{align*}
    &\nabla_{\widetilde{\mathbf{Q}}_l} \mathcal{L} := ( \nabla_{\mathbf{D}_l} \mathcal{L} ) \cdot \mathrm{cur} \mathbf{S} + \mathrm{cur} \mathbf{R}^\top \times \nabla_{\mathbf{N}_l} \mathcal{L}; \\
    &\mathrm{cur}\mathbf{R} := \mathrm{cur}\mathbf{R} - \mathbf{V}_l \times \widetilde{\mathbf{K}}_l^\top; \\
    &\mathrm{cur}\mathbf{S} := \mathrm{cur}\mathbf{S} - \widetilde{\mathbf{K}}_l; \\
    &\mathrm{grad}\mathbf{R} := \mathrm{grad}\mathbf{R} + ( \nabla_{\mathbf{N}_l} \mathcal{L} ) \times \widetilde{\mathbf{Q}}_l^\top; \\
    &\mathrm{grad}\mathbf{S} := \mathrm{grad}\mathbf{S} + ( \nabla_{\mathbf{D}_l} \mathcal{L} ) \cdot \widetilde{\mathbf{Q}}_l; \\
    &\nabla_{\mathbf{V}_l} \mathcal{L} := \mathrm{grad}\mathbf{R} \times \widetilde{\mathbf{K}}_l; \\
    &\nabla_{\widetilde{\mathbf{K}}_l} \mathcal{L} := \mathrm{grad}\mathbf{R}^\top \times \mathbf{V}_l.
\end{align*}

In practice, the described algorithm works slow when implemented in pure PyTorch, because $l$ is iterated one-by-one: \citet{transfrnn} use low-level CUDA extensions to make the algorithm practical. Instead, we propose a ``block'' version, when we iterate through blocks of $l$ of a small size $\mathcal{C}$ (we use $\mathcal{C} = 64$). In each block use explicit prefix sums on inputs of length $\mathcal{C}$ to find $\mathbf{Y}_{l:l + \mathcal{C} - 1}$, using the maintained front $\mathrm{cur}\mathbf{R}, \mathrm{cur}\mathbf{S}$. The formal algorithm is as follows. Initialize buffers $\mathrm{cur}\mathbf{R} = \mathbf{0}_{d \times M}, \mathrm{cur}\mathbf{S} = \mathbf{0}_M$. For simplicity assuming that $\mathcal{C}$ divides $L$ (extension for an opposite case is straightforward), iterate over $l = 1, \mathcal{C} + 1, \dots, L - \mathcal{C} + 1$ and compute
\begin{align}
    &\mathrm{block}\mathbf{R} := \mathrm{PS} (( \mathbf{V}_{l + l' - 1} \times \widetilde{\mathbf{K}}_{l + l' - 1}^\top)_{l' = 1}^{\mathcal{C}} ); \label{eq:ps1} \\
    &\mathrm{block}\mathbf{R} := (\mathrm{cur}\mathbf{R} + \mathrm{block}\mathbf{R}_{l'})_{l' = 1}^{\mathcal{C}}; \nonumber \\
    &\mathrm{block}\mathbf{S} := \mathrm{PS}( ( \widetilde{\mathbf{K}}_{l + l' - 1})_{l' = 1}^{\mathcal{C}} ); \label{eq:ps2} \\
    &\mathrm{block}\mathbf{S} := (\mathrm{cur}\mathbf{S} + \mathrm{block}\mathbf{S}_{l'})_{l' = 1}^\mathcal{C}; \nonumber \\
    &\mathrm{cur}\mathbf{R} := \mathrm{block}\mathbf{R}_\mathcal{C}; \nonumber \\
    &\mathrm{cur}\mathbf{S} := \mathrm{block}\mathbf{S}_\mathcal{C}; \nonumber \\
    &\mathbf{N}_{l:l + \mathcal{C} - 1} := ( \mathrm{block}\mathbf{R}_{l'} \times \widetilde{\mathbf{Q}}_{l + l' - 1} )_{l' = 1}^\mathcal{C}; \nonumber \\
    &\mathbf{D}_{l:l + \mathcal{C} - 1} := ( \mathrm{block}\mathbf{S}_{l'}^\top \times \widetilde{\mathbf{Q}}_{l + l' - 1} )_{l' = 1}^\mathcal{C}; \nonumber \\
    &\mathbf{Y}_{l:l + \mathcal{C} - 1} := ( \mathbf{N}_{l + l' - 1} / \mathbf{D}_{l + l' - 1} )_{l' = 1}^\mathcal{C}. \nonumber
\end{align}

In the ``block'' version, the number of outer sequential iterations is reduced to $L / \mathcal{C}$, resulting in $O((L / \mathcal{C}) \log \mathcal{C})$ parallel time complexity, when the logarithmic parallel algorithm is used to compute prefix sums (\ref{eq:ps1},\ref{eq:ps2}). In our experiments, we use \texttt{torch.cumsum} to compute (\ref{eq:ps1},\ref{eq:ps2}), which works fast in practice. The memory complexity of the algorithm is $O(L (d + M) + \mathcal{C} d M)$, where the second term is for storing $\mathrm{block}\mathbf{R}$. Assuming that $\mathcal{C}$ is a small constant ($\mathcal{C} = O (1)$), we conclude that the ``block'' version has $O(L (d + M) + d M)$ memory and $O(L)$ time complexity -- same as the algorithm of \citet{transfrnn}. As for hidden constants in complexity estimates, the constant inside $O(L)$ time complexity is reduced at the cost of increasing constant of the ``small'' $d M$ term in the memory complexity (when $d, M \ll L$), making the ``block'' iterative algorithm a practical choice for computing (\ref{eq:pspart}-\ref{eq:yappr}).

We further show how to back-propagate through (\ref{eq:pspart}-\ref{eq:yappr}) in $O((L / \mathcal{C}) \log \mathcal{C})$ time and $O(L (d + M) + \mathcal{C} d M)$ memory. Again, keep buffers $\mathrm{cur}\mathbf{R}, \mathrm{cur}\mathbf{S}$ as the result of forward pass, and initialize gradient buffers $\mathrm{grad}\mathbf{R} = \mathbf{0}_{d \times M}, \mathrm{grad}\mathbf{S} = \mathbf{0}_M$. Assuming that $\nabla_\mathbf{N} \mathcal{L} \in \mathbb{R}^{L \times d}, \nabla_\mathbf{D} \mathcal{L} \in \mathbb{R}^L$ are computed using automatic differentiation, iterate in a backward direction $l = L - \mathcal{C} + 1, L - 2 \mathcal{C} + 1, \dots, 1$ and compute
\begin{align*}
    &\mathrm{cur}\mathbf{R} := \mathrm{cur}\mathbf{R} - \sum_{l' = l}^{l + \mathcal{C} - 1} \mathbf{V}_{l'} \times \widetilde{\mathbf{K}}_{l'}^\top; \\
    &\mathrm{cur}\mathbf{S} := \mathrm{cur}\mathbf{S} - \sum_{l' = l}^{l + \mathcal{C} - 1} \widetilde{\mathbf{K}}_{l'}; \\
    &\mathrm{block}\mathbf{R} := \mathrm{PS} (( \mathbf{V}_{l + l' - 1} \times \widetilde{\mathbf{K}}_{l + l' - 1}^\top)_{l' = 1}^{\mathcal{C}} ); \\
    &\mathrm{block}\mathbf{R} := (\mathrm{cur}\mathbf{R} + \mathrm{block}\mathbf{R}_{l'})_{l' = 1}^{\mathcal{C}}; \\
    &\mathrm{block}\mathbf{S} := \mathrm{PS}( ( \widetilde{\mathbf{K}}_{l + l' - 1})_{l' = 1}^{\mathcal{C}} ); \\
    &\mathrm{block}\mathbf{S} := (\mathrm{cur}\mathbf{S} + \mathrm{block}\mathbf{S}_{l'})_{l' = 1}^\mathcal{C}; \\
    %%%%
    &\nabla_{\widetilde{\mathbf{Q}}_{l:l + \mathcal{C} - 1}} \mathcal{L} := ( (\nabla_{\mathbf{D}_{l + l' - 1}} \mathcal{L}) \cdot \mathrm{block} \mathbf{S}_{l'} + \mathrm{cur} \mathbf{R}_{l'}^\top \times \nabla_{\mathbf{N}_{l + l' - 1}} \mathcal{L} )_{l' = 1}^\mathcal{C}; \\
    &\mathrm{grad}\mathbf{R} := \mathrm{grad}\mathbf{R} + \sum_{l' = l}^{l + \mathcal{C} - 1} ( \nabla_{\mathbf{N}_{l'}} \mathcal{L} ) \times \widetilde{\mathbf{Q}}_{l'}^\top; \\
    &\mathrm{grad}\mathbf{S} := \mathrm{grad}\mathbf{S} + \sum_{l' = l}^{l + \mathcal{C} - 1} ( \nabla_{\mathbf{D}_{l'}} \mathcal{L}) \cdot \widetilde{\mathbf{Q}}_{l'}; \\
    %%%%
    &\mathrm{blockgrad}\mathbf{R} := \mathrm{PS} ( ( ( \nabla_{\mathbf{N}_{l + l' - 1}} \mathcal{L}) \times \widetilde{\mathbf{Q}}_{l + l' - 1}^\top )_{l' = 1}^{\mathcal{C}} ); \\
    &\mathrm{blockgrad}\mathbf{R} := (\mathrm{grad}\mathbf{R} - \mathrm{blockgrad}\mathbf{R}_{l'})_{l' = 1}^{\mathcal{C}}; \\
    &\mathrm{blockgrad}\mathbf{S} := \mathrm{PS}( ( ( \nabla_{\mathbf{D}_{l + l' - 1}} \mathcal{L} ) \cdot \widetilde{\mathbf{Q}}_{l + l' - 1} )_{l' = 1}^{\mathcal{C}} ); \\
    &\mathrm{blockgrad}\mathbf{S} := (\mathrm{grad}\mathbf{S} - \mathrm{grad}\mathbf{S}_{l'})_{l' = 1}^\mathcal{C}; \\
    %%%%
    &\nabla_{\mathbf{V}_{l:l + \mathcal{C} - 1}} \mathcal{L} := ( \mathrm{blockgrad}\mathbf{R}_{l'} \times \widetilde{\mathbf{K}}_{l + l' - 1} )_{l' = 1}^\mathcal{C}; \\
    &\nabla_{\widetilde{\mathbf{K}}_{l: l + \mathcal{C} - 1}} \mathcal{L} := ( \mathrm{blockgrad}\mathbf{R}_{l'}^\top \times \mathbf{V}_{l + l' - 1} )_{l' = 1}^\mathcal{C}.
\end{align*}

Finally, it's easy to see how to use both one-to-one and ``block'' iterative computation as part of Algorithm \ref{alg:1} to compute the update (\ref{eq:cgtr2}-\ref{eq:cgtr3}). For that, when doing a forward computation for some $n, r$, initialize $\mathrm{cur}\mathbf{R}, \mathrm{cur}\mathbf{S}$ from corresponding subvectors of $U_{B_n - 1}^{(r - 1, n - 1)}$, with the rest of the algorithm unchanged. Similarly, during a backward pass for some $n, r$, initialize $\mathrm{grad}\mathbf{R}, \mathrm{grad}\mathbf{S}$ from corresponding subvectors of $\mathcal{G}^{(n)}$ and leave the rest of the iterative back-propagation algorithm unchanged.

\section{Additional experimental details} \label{sec:addexp}

We use 15K, 30K, 100K SGD iterations in the Copying task, Penn Treebank, Enwik8 setups respectively. We use Adam optimizer \citep{adam} with $\beta_1 = 0.9, \beta_2 = 0.999$ (default configuration used in PyTorch). For the Copying task, we train with a learning rate $10^{-2}$ for 10K iterations and then decrease the learning rate to $10^{-3}$. We use a fixed learning rate of $10^{-4}$ and $2 \times 10^{-4}$ in Penn Treebank and Enwik8 experiments, respectively.

\begin{figure}
    \centering
    \includegraphics[width=0.8\textwidth]{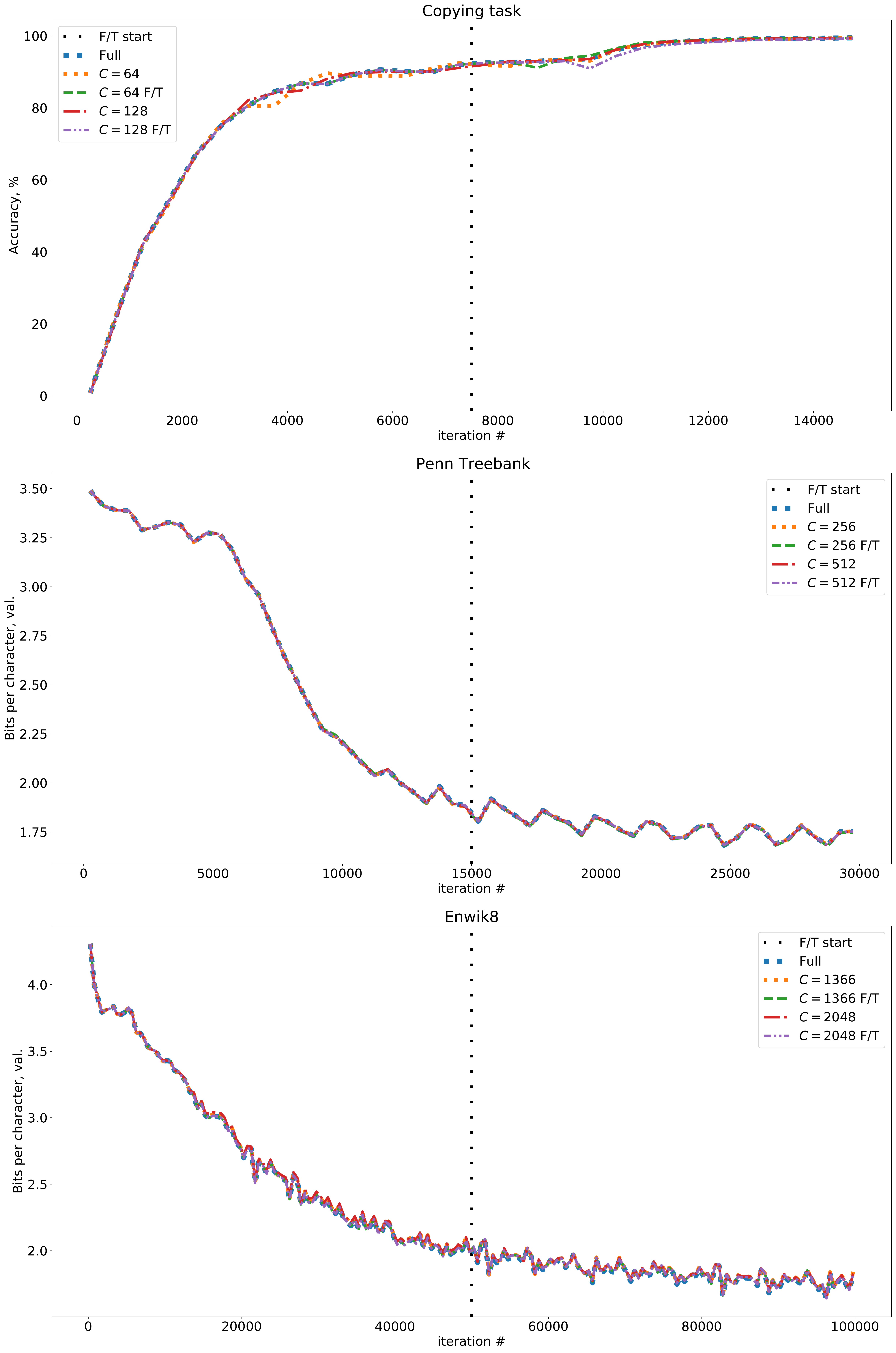}
    \caption{Bigger version of Figure \ref{fig:tr}.}
    \label{fig:app1}
\end{figure}

\begin{figure}
    \centering
    \includegraphics[width=0.8\textwidth]{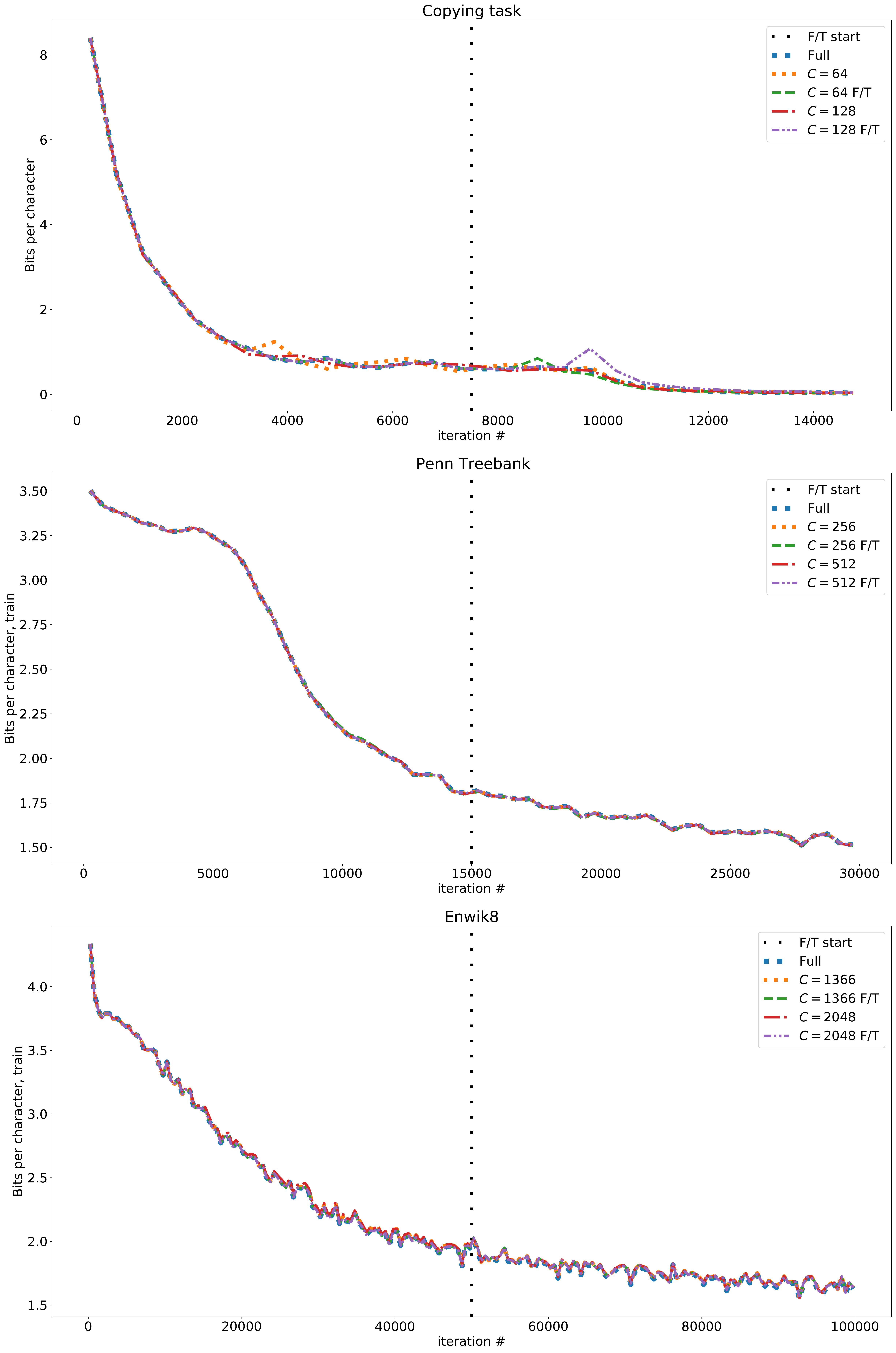}
    \caption{Bits-per-character learning curve for the Copying task and train-set learning curves for language modelling on Penn Treebank and Enwik8 respectively.}
    \label{fig:app2}
\end{figure}

Figure \ref{fig:app1} is a bigger version of Figure \ref{fig:tr} from the main text. Figure \ref{fig:app2} reports additional experimental results: bits-per-character for the Copying task and train-set learning curves for Penn Treebank and Enwik8.

\end{document}